\documentclass[letterpaper]{article} 
\usepackage{aaai24}
\usepackage{times}  
\usepackage{helvet}  
\usepackage{courier}  
\usepackage[hyphens]{url}  
\usepackage{graphicx} 
\usepackage{natbib}  
\usepackage{caption} 
\usepackage{algorithm}
\usepackage{algorithmic}

\usepackage{newfloat}
\usepackage{listings}
\DeclareCaptionStyle{ruled}{labelfont=normalfont,labelsep=colon,strut=off} 
\usepackage{adjustbox}
\usepackage{multirow}
\usepackage{amsfonts}
\usepackage{amsmath}
\usepackage{booktabs}

\usepackage{footmisc}

\floatstyle{ruled}
\newfloat{listing}{tb}{lst}{}
\floatname{listing}{Listing}
\title{Graph-Aware Contrasting for Multivariate Time-Series Classification}
\author{
    Yucheng Wang\textsuperscript{\rm 1,3}, Yuecong Xu\textsuperscript{\rm 1}, Jianfei Yang\textsuperscript{\rm 3}, \\Min Wu\textsuperscript{\rm 1}, Xiaoli Li\textsuperscript{\rm 1,2,3}, Lihua Xie\textsuperscript{\rm 3}, Zhenghua Chen\textsuperscript{\rm 1,2}\footnote{Corresponding Author}\\
}
\affiliations{
 \textsuperscript{\rm 1}Institute for Infocomm Research, A*STAR, Singapore\\
     \textsuperscript{\rm 2}Centre for Frontier AI Research, A*STAR, Singapore\\
    \textsuperscript{\rm 3}Nanyang Technological University, Singapore\\
    \{yucheng003, xuyu0014, yang0478, chen0832\}@e.ntu.edu.sg,
    \{wumin, xlli\}@i2r.a-star.edu.sg,
    elhxie@ntu.edu.sg
}

\begin{document}
\hbadness=2000000000
\vbadness=2000000000
\hfuzz=100pt

\maketitle

\begin{abstract}
Contrastive learning, as a self-supervised learning paradigm, becomes popular for Multivariate Time-Series (MTS) classification. It ensures the consistency across different views of unlabeled samples and then learns effective representations for these samples. Existing contrastive learning methods mainly focus on achieving temporal consistency with temporal augmentation and contrasting techniques, aiming to preserve temporal patterns against perturbations for MTS data. However, they overlook spatial consistency that requires the stability of individual sensors and their correlations. As MTS data typically originate from multiple sensors, ensuring spatial consistency becomes essential for the overall performance of contrastive learning on MTS data. Thus, we propose Graph-Aware Contrasting for spatial consistency across MTS data. Specifically, we propose graph augmentations including node and edge augmentations to preserve the stability of sensors and their correlations, followed by graph contrasting with both node- and graph-level contrasting to extract robust sensor- and global-level features. We further introduce multi-window temporal contrasting to ensure temporal consistency in the data for each sensor. Extensive experiments demonstrate that our proposed method achieves state-of-the-art performance on various MTS classification tasks. The code is available at https://github.com/Frank-Wang-oss/TS-GAC.
\end{abstract}


\section{Introduction}

Multivariate Time-Series (MTS) data are widely used in areas such as healthcare and industrial manufacturing for classification tasks, attracting significant research interests. To improve the performance of MTS classification, deep learning has gained popularity for learning effective representations \cite{craik2019deep, chen2021deep, deng2021graph,chen2020machine,zhao2019deep}. However, the need for substantial labeled samples poses challenges as large-scale manual labeling is impractical, limiting their applicability to real-world scenarios. To address this challenge, Contrastive Learning (CL) has emerged as a promising approach \cite{zhang2023self_arxiv, eldele2023label}. By contrasting the different views of unlabeled samples that are commonly generated by augmentations, CL enhances encoder's robustness to perturbations and learns robust and effective representations.


Researchers have recently begun exploring CL for MTS data \cite{eldele2021time,tonekaboni2021unsupervised}, with a primary focus on achieving temporal consistency by preserving temporal patterns robustly against perturbations. Specifically, temporal augmentations such as jittering or permutation are commonly used to create different views for MTS data. Encoders are then employed to extract temporal features, based on which CL is performed to make the encoders robust to temporal disturbances, thus preserving temporal patterns within MTS data. To further enhance temporal consistency, temporal contrasting is often achieved with a predictive contrastive loss when predicting the future timestamps with the past information \cite{choi2023multi,eldele2021time}.

\begin{figure}[tbp!]
    \centering\includegraphics[width = .95\linewidth]{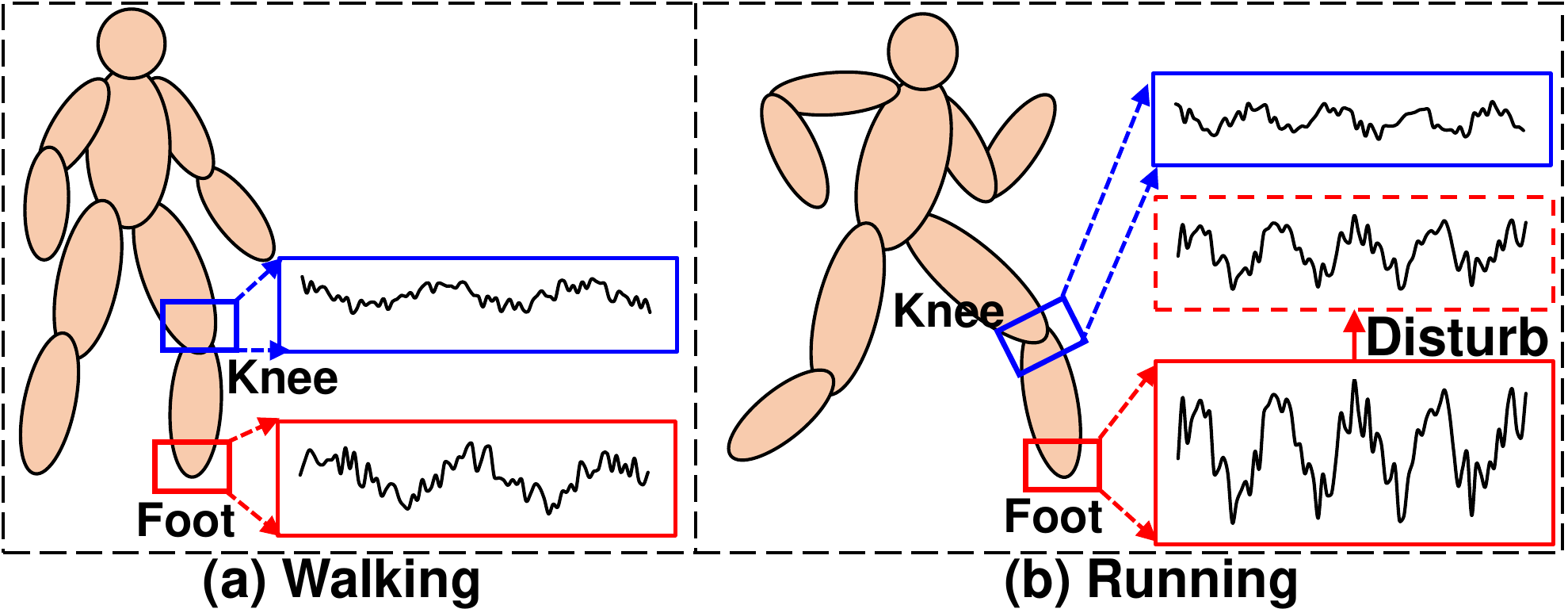}
    \caption{Signals from knee and foot for walking and running. Foot is more important for classification due to its large amplitude. (a) During walking, both knee and foot have low frequency and amplitude. (b) During running, both sensors show increased frequency and amplitude. Disturbances in the foot sensor, like insensitivity, may cause running signals to have a similar amplitude to walking signals, which may mislead a classifier and mis-classify running as walking.}
    \label{fig:sensor_level_nece}
\end{figure}

While the current methods have made progress with CL for MTS data, they mainly focused on temporal consistency while ignoring spatial consistency during the CL process. Here, the spatial consistency refers to maintaining the stability of both the individual sensors and the correlations across the different sensors. Specifically, the robustness of MTS data relies on the stability of each individual sensor, i.e., any disturbance in a sensor could have a significant impact on the classification performance of an MTS sample. We take Fig. \ref{fig:sensor_level_nece} for illustration. Amplitude disturbances, such as insensitivity, in foot signals can lead to the similar foot amplitude in walking and running actions, potentially causing a classifier to misclassify running as walking.
Thus, a robust algorithm should be able to identify disturbances within individual sensors.
Moreover, correlations exist between sensors, with certain sensors exhibiting stronger correlations across each other than with others. For example, due to the physical connection between the foot and knee, a foot sensor is more correlated with a knee sensor than a hand sensor. Preserving the robustness of these relative sensor relationships can further help learn robust sensor features \cite{yu2017spatio,jia2020graphsleepnet}. As MTS data typically originate from multiple sensors, it is crucial to ensure the spatial consistency to enhance the overall CL performance on MTS data.


The above discussion motivates us to propose a novel approach called Graph-Aware Contrasting for MTS data (TS-GAC). To achieve spatial consistency, specific augmentation and contrasting methods tailored for MTS data are designed. We first design graph augmentations, involving node and edge augmentations, to augment MTS data. For node augmentations, we apply temporal and frequency augmentations \cite{zhang2022self,yang2022unsupervised} to fully augment each sensor, while edge augmentations are designed to augment sensor correlations, ensuring robustness in the relationships between sensors. By capturing the augmented sensor correlations, Graph Neural Network (GNN)~\cite{wang2023multivariate,jia2020graphsleepnet} is utilized to update sensor features.

With updated sensor features, we then design graph contrasting which incorporates both node- and graph-level contrasting to learn robust sensor- and global-level features. For node-level contrasting, we create two views using the proposed augmentations and contrast the sensors in different views within each MTS sample to ensure the robustness of each sensor against perturbations. Additionally, we map the sensor features to global features and introduce graph-level contrasting by contrasting MTS samples in different views within each training batch. 
Furthermore, we achieve temporal consistency for each sensor through temporal contrasting by following prior works \cite{choi2023multi,eldele2021time}. Due to the dynamic nature of sensor correlations in MTS data \cite{wang2023multivariate}, we propose segmenting a sample into multiple windows, enabling us to incorporate multi-window temporal contrasting which ensures the consistency of temporal patterns within each sensor.

In summary, our contributions are three folds. First, to promote spatial consistency, we propose novel graph augmentations to enhance the quality of augmented views for MTS data. The graph augmentations involve node and edge augmentations, aiming to augment sensors and their correlations respectively. Second, we design graph contrasting that includes node- and graph-level contrasting, facilitating the learning of robust sensor- and global-level features. We also introduce a multi-window temporal contrasting to achieve temporal consistency for each sensor. Third, we conduct extensive experiments on ten public MTS datasets, showing that our TS-GAC achieves state-of-the-art performance.

\section{Related Work}
\paragraph{Contrastive Learning (CL)}
As a self-supervised learning paradigm, CL has gained popularity due to its ability to learn effective features from unlabeled samples by bringing positive pairs closer while pushing negative pairs farther \cite{zhang2023self_arxiv, eldele2023label}. Augmentations are commonly used to create positive pairs, generating augmented samples from different perspectives. Negative pairs, on the other hand, are created using the remaining samples in the same batch \cite{chen2020simple} or stored in a memory bank \cite{he2020momentum}. Contrasting these positive and negative pairs helps encoders become robust to perturbations, ensuring consistency in the learned features, and thus learning robust and effective features from unlabeled data.

Researchers have proven the effectiveness of CL in image tasks \cite{hjelm2018learning,he2020momentum,caron2020unsupervised,chen2020simple}. MoCo~\cite{he2020momentum} designed a momentum encoder with a memory bank to store negative samples, achieving desirable performance with limited computational resources. SimCLR~\cite{chen2020simple} adopted larger batches of negative pairs and achieved comparable results to supervised learning. Inspired by SimCLR, MoCo-v2 \cite{chen2020improved} improved performance with powerful augmentations without requiring large batches. Besides, negative pairs may occupy computation resources, so BYOL \cite{grill2020bootstrap} and SimSiam \cite{chen2021exploring} learned representations with only positive pairs. Although these methods have achieved decent performance, they are proposed for images. Different from images, MTS data contain both temporal and spatial information from multiple sensors, making traditional image-based augmentation and contrasting methods unsuitable for MTS data.

\paragraph{CL for MTS Data}
Pioneering works have successfully utilized CL techniques to learn decent representations from unlabeled MTS data, primarily focusing on achieving temporal consistency \cite{poppelbaum2022contrastive,khaertdinov2021contrastive,hao2023micos,yue2022ts2vec,eldele2021time}. Specifically, they augmented MTS data with temporal augmentations such as jittering, cropping, and sub-series, and then conducted CL to ensure encoders robustness to temporal disturbances. Meanwhile, some works \cite{choi2023multi,eldele2021time} also introduced temporal contrasting by summarizing past information for contrasting with future timestamps, further enforcing robustness to perturbations within timestamps. 

While these works advanced CL for MTS data by ensuring temporal consistency, they overlook spatial consistency for MTS data. Some recent works proposed to incorporate spatial information, e.g., sensor correlations, into CL frameworks. For example, TAGCN \cite{zhang2023exploring} utilized GNN to extract features from sub-series of MTS data and then performed CL. Additionally, TSGCC \cite{zhang4474418graph} designed a graph-based method to compute weights between samples for clustering by instance- and clustering- contrasting. However, these methods only utilized GNN to extract spatial information within MTS data, while still overlooking spatial consistency to achieve better CL for MTS data. Although a few recent studies \cite{chen2022cass,li2022SPGCL} explored channel-wise signal augmentations, graph-level augmentations and contrasting are still under-explored, limiting their ability to achieve robust spatial consistency for MTS data.

To overcome the limitations, we propose TS-GAC which incorporates both the graph augmentation and graph contrasting techniques to ensure spatial consistency during the CL process for MTS classification.



\section{Methodology}
\renewcommand{\dblfloatpagefraction}{.9}
\begin{figure*}[htbp!]
    \centering
    \includegraphics[width = .9\linewidth]{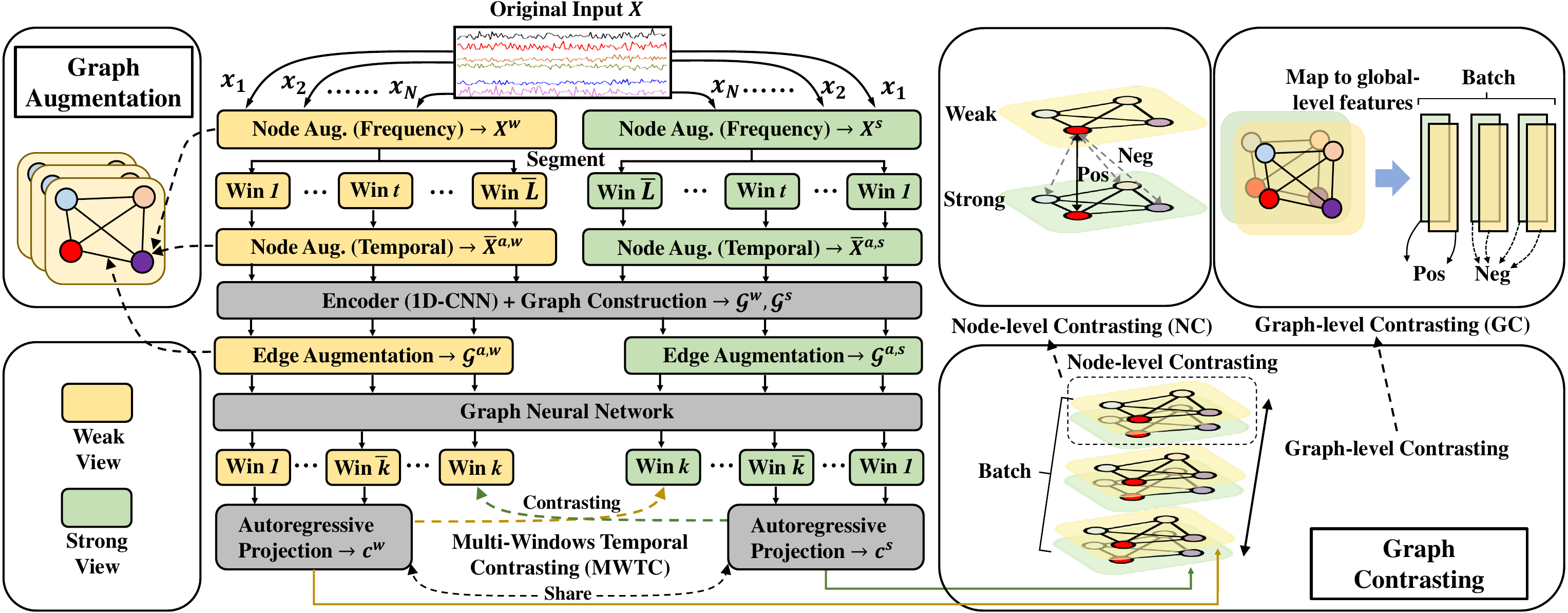}
    \caption{Overall structure of TS-GAC. (1) Graph augmentations to augment MTS data effectively, generating weak and strong views. The graph augmentations involve node and edge augmentations, where node augmentations include both frequency and temporal augmentations to fully augment sensors. Node frequency augmentations are first applied, followed by segmenting augmented samples into multiple windows by considering the dynamic local patterns in MTS data. Node temporal augmentations are utilized within each window, followed by 1D-CNN to process these windows. Subsequently, graphs are constructed and augmented through edge augmentations, and then processed by GNN. (2) Graph contrasting includes NC and GC to achieve spatial consistency. NC ensures robust sensors by pulling closer corresponding sensors in different views and pushing father different sensors in those views within each sample. GC ensures robust global features by pulling closer corresponding samples in different views and pushing father different samples in those views within each batch. MWTC further achieves temporal consistency for each sensor by summarizing past windows to contrast with future windows in another view.}
    \label{fig:overall}
\end{figure*}

\subsection{Problem Formulation}
Given a dataset with $n$ unlabeled MTS samples $\mathcal{X} = \{X_j\}_{j=1}^n$, each sample $X_j\in\mathbb{R}^{N\times{L}}$ is collected from $N$ sensors with $T$ timestamps. Our objective is to perform contrastive learning scheme that can achieve spatial consistency for MTS data, enabling the training of an encoder $\mathcal{F}$ without relying on labels. This approach allows us to achieve enhanced CL performance and thus extract effective representations $h_j = \mathcal{F}(X_j)\in\mathbb{R}^{d}$. With $h_j$, we employ a simple classifier, e.g., a multi-layer perceptron, to obtain class probabilities $y_j\in\mathbb{R}^{c}$, where $c$ represents the number of classes in the classification task. For simplicity, the subscript $j$ is removed, and we denote an MTS sample as $X$.

\subsection{Overall Structure}
Fig.~\ref{fig:overall} shows the overall structure of TS-GAC, which aims to achieve spatial consistency in CL for MTS classification. Specific augmentation and contrasting techniques are tailored for MTS data. For augmentation, we consider node and edge augmentations to augment individual sensors and their correlations, generating weak and strong views for each sample. Node frequency augmentations are applied first, followed by segmenting augmented samples into multiple windows considering the dynamic local patterns in MTS data. Node temporal augmentations are utilized within each window, followed by a 1-Dimensional Convolutional Neural Network (1D-CNN) to process these windows. Subsequently, graphs are constructed with each sensor as a node and sensor correlations are edges. The constructed graphs are further augmented by edge augmentations, and then processed by a GNN-based encoder to learn representations. Next, to achieve spatial consistency, we design graph contrasting including Node-level Contrasting (NC) and Graph-level Contrasting (GC). NC enables the contrasting of sensors within each sample to learn robust sensor-level features while GC allows the contrasting of samples within each training batch, promoting the learning of robust global-level features. We further introduce Multi-Window Temporal Contrasting (MWTC) to ensure temporal consistency for each sensor, by utilizing past windows in one view to predict the future windows in another view.



\subsection{Augmentation}
CL learns robust representations by contrasting different views of unlabeled data, which are commonly generated by augmentations. Then, the augmented views from the same data are pulled closer and the views from different data are simultaneously pushed farther for representation learning. Thus, augmentations are critical for CL to learn robust and generalizable representations. To enhance augmentation quality for MTS data, we consider its multi-source nature, i.e., collected from multiple sensors \cite{zhao2019deep}. We argue that augmentations for MTS data should be able to ensure the learning of robust sensor features and sensor correlations. For this purpose, we design node and edge augmentations that augment individual sensors and their correlations respectively. Further, following \cite{eldele2021time}, we generate weak and strong views, i.e., weakly and strongly augmented, for each sample with the augmentations for subsequent contrasting.


\subsubsection{Node Augmentations}

We perform both frequency and temporal augmentations for the nodes (i.e., sensors).

\textit{\textbf{Frequency augmentations}}: We utilize frequency augmentations to augment individual sensors, as the augmentations are widely recognized as effective in augmenting time-series data \cite{zhang2022self,zhang2023self}. This involves transforming the signals of each sensor into the frequency domain and augmenting the extracted frequency features. The augmented frequency features are then transformed back into the temporal domain to obtain augmented signals.

Particularly, we adopt Discrete Wavelet Transform (DWT) \cite{boggess2002first} to decompose signals into detail and approximation coefficients using high-pass and low-pass filters, representing detailed and general trends within the signals, respectively. To generate weak and strong views, we add Gaussian noise to the detail and approximation coefficients respectively. The augmented frequency features are then transformed back into the temporal domain using inverse DWT (iDWT) to obtain the augmented signals. Mathematically, frequency augmentations are achieved via Eq.~(\ref{eq:dwt}), where $\eta_{A, i}$ and $\eta_{D, i}$ denote the approximation and detail coefficients for the $i$-th sensor, and $\xi$ represents the noise added to coefficients. We denote $\{X^w, X^s\}$ as the augmented signals in weak and strong views.
\begin{gather}
        \eta_{A, i}, \eta_{D,i} = DWT(x_i),\nonumber\\
                    \label{eq:dwt}
        \eta^s_{A, i} = \eta_{A, i} + \xi,
        \eta^w_{D, i} = \eta_{D, i} + \xi,\\
        x_i^s = iDWT(\eta^s_{A, i}, \eta_{D,i}),
        x_i^w = iDWT(\eta_{A, i}, \eta^w_{D,i})\nonumber.
\end{gather}

\begin{figure}[htbp!]
    \centering
    \includegraphics[width = 1.\linewidth]{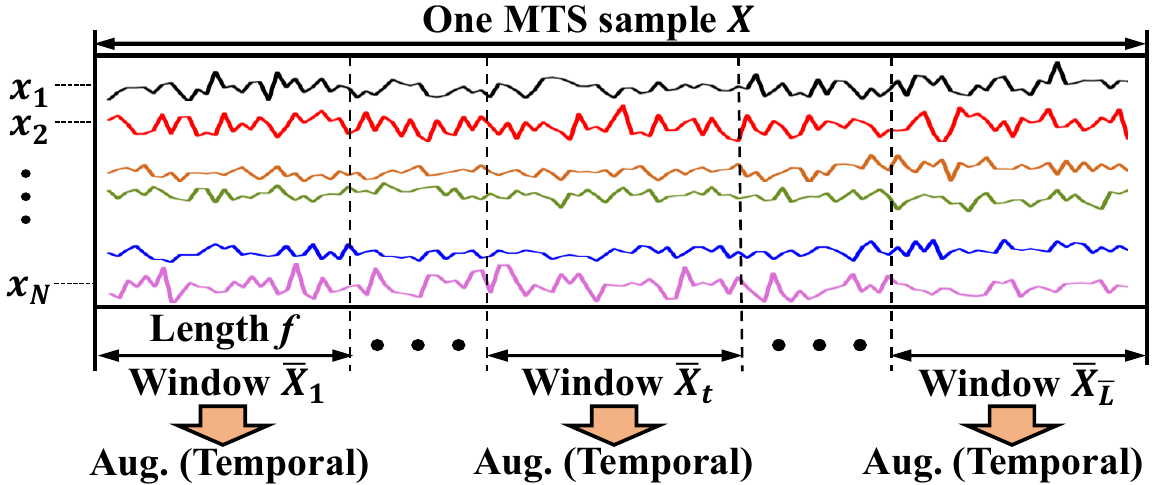}
    \caption{The multi-window segmentation to generate multiple windows for one MTS sample.}
    \label{fig:windows}
\end{figure}

\textit{\textbf{Temporal augmentations}}: We further introduce temporal augmentations to augment each sensor due to its importance in augmenting time-series data \cite{poppelbaum2022contrastive,khaertdinov2021contrastive}. Before temporal augmentations, we note that MTS data show dynamic properties, i.e., local patterns of MTS data are dynamically changing \cite{wang2023multivariate}. To capture such properties, we segment each MTS sample into mini windows. As displayed in Fig.~\ref{fig:windows}, given the window with length $f$, we segment an MTS sample into $\Bar{L} = [L/f]$ windows, where $[\,]$ represents truncation. Thus, we obtain $X^w = \{\Bar{X}^w_t\}_{t=1}^{\Bar{L}}$ for the weak view, where $t$ is the index of the window, and $\Bar{X}^w_t = \{\Bar{x}^w_{t, i}\}_{i=1}^N\in\mathbb{R}^{N\times{f}}$ contains the local patterns, including local sensor features and correlations. The windows in the strong view $\{\Bar{X}^s_t\}_{t=1}^{\Bar{L}}$ are obtained in the same way. In this case, if we conduct temporal augmentations before segmentation, it is hard to augment each window averagely, so we propose augmenting each window after segmentation. 

We adopt permutation for temporal augmentations due to its wide application \cite{eldele2021time,poppelbaum2022contrastive} and augment each sensor of each window. After augmentation, we obtain the augmented windows, e.g., $\{\Bar{X}^{a,w}_t\}_{t=1}^{\Bar{L}}$ in the weak view, where $\Bar{X}^{a,w}_t = \{\Bar{x}_{t, i}^{a,w}\}_{i=1}^N$. 1D-CNN is then utilized as an encoder to capture the temporal information between windows \cite{jin2022position}, \textit{whose details are attached in our supplementary materials}. With the encoder, we learn updated windows, e.g., $\{{Z}^w_t\}_{t=1}^{k}$ for the weak view, where ${Z}^w_{t} = \{{z}_{t, i}^w\}_{i=1}^N$. Similar notations such as $\Bar{X}^{a,s}_t$ and ${Z}^s_{t}$ apply to the strong view.

\subsubsection{Edge Augmentations}
The correlations between sensors should remain robust due to their importance for learning sensor features~\cite{jia2020graphsleepnet,ijcai2022p332}. 
To ensure robust sensor relationships, we begin by constructing graphs whose nodes and edges represent sensors and the correlations between these sensors respectively. Augmenting the edges allows us to augment the relations effectively.
For graph construction, we note that correlated sensors should follow similar properties and their features should be similar in the feature space, so we leverage the features similarities to define the sensor correlations. Given the features ${Z}_t = \{{z}_{t, i}\}_{i=1}^N\in\mathbb{R}^{N\times{f}}$, we compute the correlation between sensors $i$ and $j$ using the dot product of their features, i.e., $e_{t, ij} = {z}_{t, i}({z}_{t, j})^T$. Then, the softmax function is used to restrict the correlations within the range [0,1]. Multiple graphs are built based on the windows for two views. For the weak view, the graph for $t^{th}$ window is denoted as $\mathcal{G}_t^{w} = ({Z}_t^{w}, E_t^{w})$, where $E_t^{w} = \{e_{t, ij}^{w}\}_{i, j}^N$. Similar graphs $\mathcal{G}_t^{s}$ are obtained for the strong view.

We then introduce edge augmentations to augment the correlations between sensors. A naive approach would be randomly adding noise, replacing, or dropping certain edges for graph augmentation \cite{you2020graph}. However, this method may introduce excessive bias and significantly alter the topological structure within MTS data. Note that GNN updates sensor features based on their correlations with other sensors. Thus, strong correlations ensure more information propagation, making them more crucial than weak correlations. Randomly disturbing these strong correlations can introduce excessive bias. To address this issue, it is necessary to add constraints for the edge augmentation. Thus, we propose retaining the $s$ strongest correlations (i.e., top-$s$ correlations) for each sensor and augmenting the remaining correlations by replacing them with random values within the range [0, 1]. This approach allows us to fully augment sensor correlations while preserving the topological information within MTS data as much as possible. Specifically, we retain more strong correlations for graphs in the weak view and fewer strong correlations for graphs in the strong view. The resulting augmented graph for the $t^{th}$ window in the weak view is denoted as ${\mathcal{G}}_t^{a,w} = ({Z}_t^{w}, {E}_t^{a,w})$, and ${E}_t^{a,w}$ are augmented sensor correlations. Similarly, ${\mathcal{G}}_t^{a,s}$ denotes the augmented graph for the strong view.

With the augmented graphs, we adopt GNN to update sensor features by leveraging the augmented correlations as conventional works did \cite{jia2020graphsleepnet,wang2023multivariate}.
Particularly, the features for sensor $i$ in the weak view are updated by a nonlinear function, i.e., $z_{t,i}^{w} = \sigma(\sum_{j}^N{z}_{t,j}^{w}{e}^{a,w}_{t, ij}W_g)$, where $W_g$ are learnable weights. The updated sensor features $z_{t,i}^{w}$ and $z_{t,i}^{s}$ in weak and strong views are then used for subsequent contrasting.

\subsection{Contrasting}
With the augmentations to generate weak and strong views, we design graph contrasting to achieve spatial consistency and further design MWTC to achieve temporal consistency for each sensor. We begin by presenting MWTC in this section, as it learns high-level sensor features within multi-window for subsequent graph contrasting.

\subsubsection{Multi-Window Temporal Contrasting}
MWTC operates at the sensor-level, ensuring temporal consistency for each sensor. It is noted that the multi-window of each sensor show temporal dependencies where future windows are normally affected and dependent on past windows, which can be incorporated to keep the multi-window robust. Inspired by the idea of predictive coding \cite{oord2018representation} and temporal contrasting \cite{choi2023multi,eldele2021time}, we propose to summarize past windows in one view to contrast with the future windows in another view. By doing so, we aim to maintain the temporal dependency robustness against perturbations to the windows, enabling that the temporal patterns within MTS data are preserved.

Specifically, we introduce an auto-regressive model $f_a$ to summarize the sensor features in past $\Bar{k}$ windows, i.e., $c_i^{w} = f_a(z_{1,i}^{w}, ..., z_{\Bar{k},i}^{w}|W_a)$, representing the summarized vectors for the $i$-th sensor in the weak view. $c_i^{w}$ is then to predict future windows, i.e., $\Bar{z}_{\Bar{k}+1, i}^{w} = f_{\Bar{k}+1}(c_i^{w}), ..., \Bar{z}_{k, i}^{w} = f_{k}(c_i^{w})$, where $f_{\Bar{k}+1}(\cdot),..., f_{k}(\cdot)$ are nonlinear functions to predict the $(\Bar{k}+1)$-th, ..., $k$-th windows. Similar operations are conducted for the strong view. Here, we adopt a transformer model for $f_a$ following~\cite{eldele2021time}, the detail of which is attached in our supplementary materials. $\mathcal{L}^{s\to{w}}_{MWTC}$ in Eq. (\ref{eq:mulwinTC}) is the loss using the past windows in the strong view to predict the future windows in the weak view. Here, the predicted window $\Bar{z}_{t,i}^{s}$ should exhibit similarity with its positive pair ${z}_{t,i}^{w}$, while being dissimilar with its negative pairs ${z}_{v,i}^{w}, v\in\hat{\mathcal{V}}_{t,i}$, where $\hat{\mathcal{V}}_{t,i}$ denotes the set of windows excluding the $t$-th window for sensor $i$.


\begin{equation}
    \label{eq:mulwinTC}
    \mathcal{L}^{s\to{w}}_{MWTC} = \frac{-1}{N(k-\Bar{k})}\sum_i^N\sum_{t=\Bar{k}}^klog\frac{exp((\Bar{z}_{t,i}^{s})^T{z}_{t,i}^{w})}{\sum_{v\in\hat{\mathcal{V}}_{t,i}}exp((\Bar{z}_{t,i}^{s})^T{z}_{v,i}^{w})}.
\end{equation}
Similarly, we can obtain $\mathcal{L}^{w\to{s}}_{MWTC}$ and thus obtain $\mathcal{L}_{MWTC} = \mathcal{L}^{s\to{w}}_{MWTC} + \mathcal{L}^{w\to{s}}_{MWTC}$ for sample $X$.

\subsubsection{Graph Contrasting}

We propose graph contrasting to achieve spatial consistency, including Node-level Contrasting (NC) and Graph-level Contrasting (GC) to learn robust sensor- and global-level features. NC is achieved by contrasting sensors in different views within each MTS sample while GC is achieved by contrasting the samples within each training batch. Notably, we leverage the vectors $\{c_i\}_{i=1}^N$ for graph contrasting, as the vectors represent the high-level features by summarizing the sensor-level features within multi-window. By utilizing the high-level features, we can achieve more effective graph contrasting.

\textit{\textbf{Node-level Contrasting}}:
NC is designed to learn robust sensor-level features. Specifically, it aims to maximize the similarity between the corresponding sensors in two views while minimizing the similarity between different sensors in those views. By doing so, NC encourages the encoder to learn features against perturbations to each sensor. Eq.~(\ref{eq:nodects}) presents the node-level contrastive loss, where $\hat{\mathcal{V}}_i$ denotes the set of sensors excluding sensor $i$. The visualization process is shown in NC of Fig.~\ref{fig:overall}.
\begin{equation}
    \label{eq:nodects}
        \mathcal{L}_{NC}^{s\to{w}} = -\frac{1}{N}\sum_i^Nlog\frac{exp(f_{sim}(c_i^s, c_i^w)/\tau)}{\sum_{v\in\hat{\mathcal{V}}_i}exp(f_{sim}(c_i^s, c_v^w)/\tau)}.
\end{equation}
Here $f_{sim}(a,b)$ is a function to measure the similarity of samples implemented as the dot product $a^Tb$, and $\tau$ is a temperature parameter. $\mathcal{L}_{NC}^{s\to{w}}$ denotes that the sensors in the strong view are contrasted with the positive and negative pairs in the weak view. Similarly, we can obtain $\mathcal{L}_{NC}^{w\to{s}}$ and thus obtain $\mathcal{L}_{NC} = \mathcal{L}_{NC}^{s\to{w}} + \mathcal{L}_{NC}^{w\to{s}}$ for sample $X$.

\textit{\textbf{Graph-level Contrasting}}:
GC aims to learn robust global-level features by contrasting samples within each training batch. For subsequent contrasting, we here obtain the global-level features by stacking all sensor features. For the weak view, $g^{w} = [c_1^{w}|...|c_N^{w}]$, where $[\,]$ denotes concatenation. Similar operations are conducted for the strong view.

To learn robust global-level features, GC is achieved by maximizing the similarity between the corresponding samples in two views and simultaneously minimizing the similarity between the different samples in those views. Given a batch of $B$ MTS samples, we have 2$B$ augmented samples from two augmented views. The corresponding samples in two views are treated as positive pairs, and each view of the sample can form 2$B$-2 negative pairs with the remaining augmented samples. We denote the global-level features of the $p$-th augmented samples in weak and strong views within the batch as $g^{\{w,s\}}_p$. Accordingly, the graph-level contrasting is demonstrated as Eq.~(\ref{eq:graphcts}), which denotes that the samples in the strong view are contrasted with the remaining augmented samples in the batch. Here, $\hat{\mathcal{V}}_p$ denotes the set of samples in the batch excluding the $p$-th sample.
\begin{equation}
    \label{eq:graphcts}
        \mathcal{L}_{GC}^{s} = -\frac{1}{B}\sum_{p=1}^Blog\frac{exp(f_{sim}(g^{s}_p,g^w_p)/\tau)}{\sum_{v\in\hat{\mathcal{V}}_{p}}exp(f_{sim}(g^{s}_p,g^{\{w,s\}}_v)/\tau}.
\end{equation}
Similarly, we can obtain $\mathcal{L}_{GC}^{w}$ for the weak view and thus obtain $\mathcal{L}_{GC} = \mathcal{L}_{GC}^{s} + \mathcal{L}_{GC}^{w}$. 

Finally, we combine MWTC, NC, and GC to form the final self-supervised loss as Eq.~(\ref{eq:overaloss}), where $\lambda_{MWTC}$, $\lambda_{NC}$, and $\lambda_{GC}$ are hyperparameters that denote relative weights of the losses. Notably, MWTC and NC are both achieved for each MTS sample, so they are denoted as $\mathcal{L}_{p,MWTC}$ and $\mathcal{L}_{p,NC}$ for the $p$-th sample.
\begin{equation}
    \label{eq:overaloss}
    \mathcal{L} = \lambda_{MWTC}\sum_p^B\mathcal{L}_{p,MWTC} + \lambda_{NC}\sum_p^B\mathcal{L}_{p,NC} + \lambda_{GC}\mathcal{L}_{GC}.
\end{equation}

\section{Experimental Results}
\paragraph{Datasets}
We examine our method on ten public MTS datasets for classification, including Human Activity Recognition (HAR)~\cite{anguita2012human}, ISRUC~\cite{khalighi2016isruc}, and eight large datasets from UEA archive, i.e., ArticularyWordRecognition (AWR), FingerMovements (FM), SpokenArabicDigitsEq (SAD), CharacterTrajectories (CT), FaceDetection (FD), InsectWingbeat (IW), MotorImagery (MI), and SelfRegulationSCP1 (SRSCP1). For HAR and ISRUC, we randomly split them into 80\% and 20\% for training and testing, while for those from UEA archive, we directly adopt their pre-defined train-test splits. \textit{The statistics of the datasets are in the Appendix}. 

\paragraph{Evaluation}
For evaluation, we follow the standard linear classification scheme as current methods did~\cite{eldele2021time,yue2022ts2vec}, i.e., train an encoder with only training data in a self-supervised manner and then train a linear classifier on top of the pre-trained encoder. To evaluate performance, we adopt two metrics, i.e., Accuracy (Accu.) and Macro-averaged F1-Score (MF1) \cite{eldele2021time,meng2022mhccl}. Besides, to reduce the effect of random initialization, we conduct ten times for all experiments and take the average results for comparisons. The standard variations are reported to show the robustness of the results.

\paragraph{Implementation Details}
All methods are conducted with NVIDIA GeForce RTX 3080Ti and implemented by PyTorch \cite{paszke2019pytorch}. We set the batch size as 128 and choose ADAM as the optimizer with a learning rate of 3e-4. We pre-train the model and train the linear classifier 40 epochs. \textit{More implementation details are in the Appendix}.

\subsection{Comparisons with State-of-the-Arts}

We compare our method with SOTA methods, including SimCLR~\cite{chen2020simple}, TNC~\cite{tonekaboni2021unsupervised}, TS-TCC~\cite{eldele2021time}, TS2Vec~\cite{yue2022ts2vec}, MHCCL~\cite{meng2022mhccl}, CaSS \cite{chen2022cass}, and TAGCN \cite{zhang2023exploring}. All methods are re-implemented based on their original settings except for the encoders, which are replaced by the same encoder as ours for fair comparisons.

Table~\ref{tab:sota} shows the comparisons with SOTA methods. From the table, we observe that TS-GAC achieves the best performance on eight out of ten datasets. Particularly, TS-GAC gains great improvements on HAR and ISRUC, improving by 1.44\% and 3.13\% respectively regarding accuracy. In the remaining cases where TS-GAC achieves the second best, the gaps of TS-GAC with the best result are marginal, e.g., only 0.4\% lower than the best accuracy for FM. Meanwhile, TS-GAC has smaller variances, indicating that our TS-GAC is more robust and stable.

\begin{figure}[htbp!]
    \centering
    \includegraphics[width = 1.\linewidth]{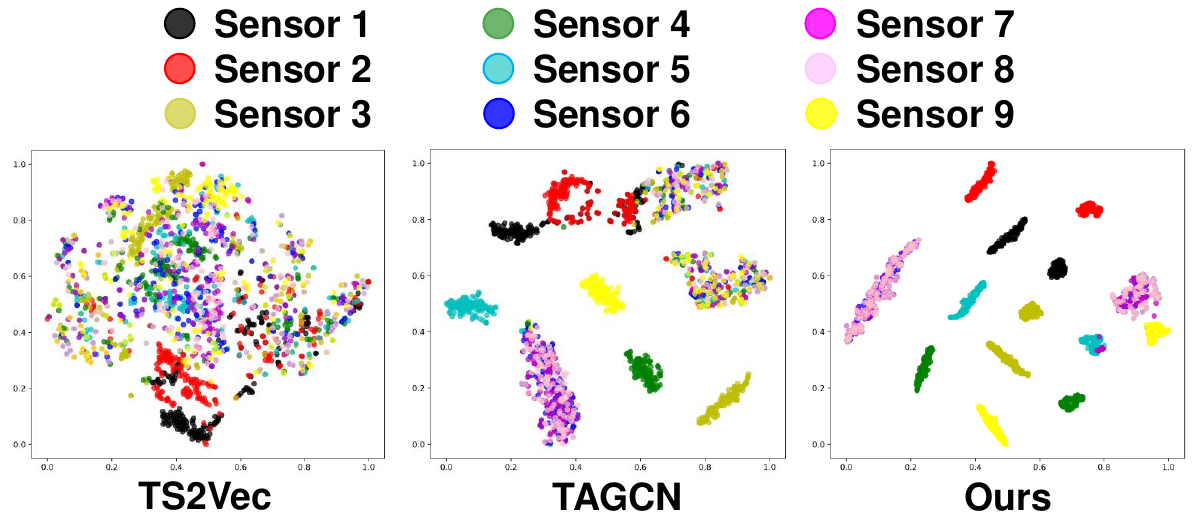}
    \caption{Visualization for sensor features.}
    \label{fig:visual_sensor}
\end{figure}
\begin{figure}[htbp!]
    \centering
    \includegraphics[width = 1\linewidth]{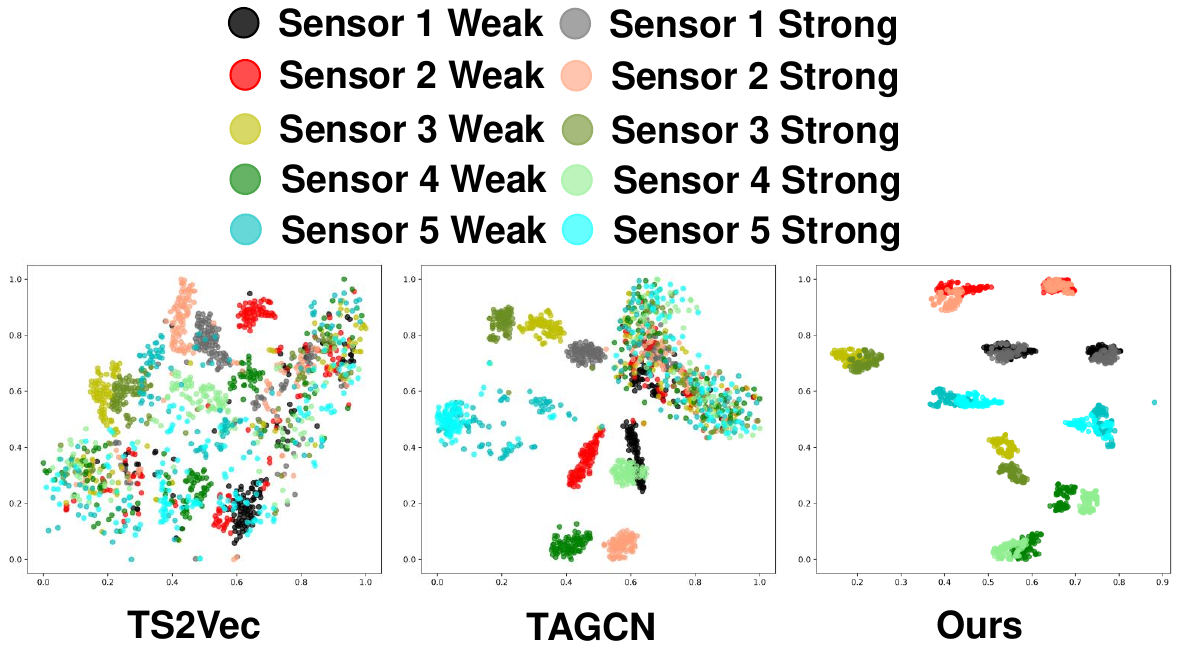}
    \caption{Visualization for spatial consistency.}
    \label{fig:visual_spatial}
\end{figure}

The superior performance can be attributed to the spatial consistency achieved by TS-GAC. To intuitively demonstrate this spatial consistency, we visualized sensor features from different views, comparing TS-GAC with two competitive methods, TS2Vec and TAGCN. We first visualized the individual sensor features. As shown in Fig. \ref{fig:visual_sensor}, TS-GAC exhibits clearer sensor clusters than TS2Vec and TAGCN, emphasizing its ability to learn robust sensor features.
Based on the clear sensor features, the features extracted from weak and strong views are aligned. Specifically, TS-GAC obtains closer feature clusters for the same sensors in weak and strong views, demonstrating its capability to learn consistent sensor features across different perspectives.

\begin{table*}[htbp]
  \centering
  \small
    \begin{tabular}{c|c|ccccccc|c}
    \toprule
    \toprule
    Datasets & Metrics & SimCLR   & TNC   & TS-TCC & TS2Vec & MHCCL &CaSS & TAGCN & TS-GAC (Ours) \\
    \midrule
    \multirow{2}[2]{*}{HAR} & Accu  & 89.97$\pm$0.46 &  81.10$\pm$1.88 & 91.66$\pm$0.42 & {92.78$\pm$0.32} & 82.95$\pm$0.55& 82.64$\pm$0.31& \underline{92.83$\pm$0.28} & \textbf{94.27$\pm$0.12} \\
          & MF1   & 89.91$\pm$0.42 &  78.24$\pm$2.91 & 91.86$\pm$0.40 & \underline{92.78$\pm$0.33} & 82.70$\pm$0.62& 82.34$\pm$0.31& 92.66$\pm$0.29 & \textbf{94.07$\pm$0.14} \\
              \midrule
    \multirow{2}[2]{*}{ISRUC} & Accu  & 75.07$\pm$0.40 &  77.69$\pm$1.28 & {80.50$\pm$0.42} & 76.32$\pm$0.48 & 74.71$\pm$0.98& \underline{81.09$\pm$0.19}& 77.21$\pm$0.21 & \textbf{84.22$\pm$0.17} \\
          & MF1   & 72.60$\pm$0.38 &  64.08$\pm$1.60 & {79.12$\pm$0.40} & 74.44$\pm$0.59 & 72.09$\pm$1.23& \underline{79.73$\pm$0.29}& 76.23$\pm$0.27 & \textbf{83.45$\pm$0.23} \\
    \midrule
    \multirow{2}[2]{*}{AWR} & Accu  & 92.78$\pm$0.80 &  82.60$\pm$4.21 & 89.44$\pm$0.68 & \underline{98.30$\pm$0.09} & 93.00$\pm$0.56& 97.47$\pm$0.16& 97.87$\pm$0.27 & \textbf{98.33$\pm$0.08} \\
          & MF1   & 92.69$\pm$0.82 &  77.42$\pm$5.34 & 89.51$\pm$0.73 & \underline{98.29$\pm$0.10} & 93.14$\pm$0.75& 97.46$\pm$0.16& 97.86$\pm$0.27 & \textbf{98.33$\pm$0.07} \\
    \midrule
    \multirow{2}[2]{*}{FM} & Accu  & 50.52$\pm$2.04 &  48.90$\pm$2.42 & 47.40$\pm$1.63 & 47.10$\pm$4.22 & \textbf{52.40$\pm$2.28}& 50.00$\pm$1.79& 51.50$\pm$1.91 & \underline{52.00$\pm$1.54} \\
          & MF1   & {47.35$\pm$1.97} &  43.02$\pm$5.25 & 47.36$\pm$1.64 & 47.03$\pm$4.18 & \textbf{49.82$\pm$3.06}& 35.10$\pm$2.01& \underline{49.52$\pm$2.04} & 48.78$\pm$0.71 \\
    \midrule
    \multirow{2}[2]{*}{SAD} & Accu  & 93.72$\pm$0.50 &  90.30$\pm$1.36 & 95.20$\pm$0.15 & {97.31$\pm$0.19} & 95.91$\pm$0.56& 97.44$\pm$0.07& \underline{97.50$\pm$0.03} & \textbf{97.99$\pm$0.05} \\
          & MF1   & 93.76$\pm$0.50 &  88.83$\pm$1.42 & 95.24$\pm$0.15 & {97.31$\pm$0.19} & 95.92$\pm$0.45& 97.45$\pm$0.07& \underline{97.52$\pm$0.03} & \textbf{97.99$\pm$0.05} \\
        \midrule
        \multirow{2}[2]{*}{CT} & Accu  & 96.14$\pm$0.20 &  96.23$\pm$1.24 & 98.61$\pm$0.17 & 98.68$\pm$0.02 & 98.21$\pm$0.10 & 97.16$\pm$0.07 & \textbf{98.89$\pm$0.10} & \underline{98.82$\pm$0.05} \\
          & MF1   & 95.86$\pm$0.24 &  95.98$\pm$1.56 & 98.49$\pm$0.19 & 98.59$\pm$0.02 & 95.62$\pm$0.11 & 96.92$\pm$0.08 & \textbf{98.81$\pm$0.09} & \underline{98.77$\pm$0.06} \\
    \midrule
    \multirow{2}[2]{*}{FD} & Accu  & 59.18$\pm$1.23 &  50.15$\pm$0.61 & 58.00$\pm$1.71 & \underline{59.60$\pm$0.61} & 55.26$\pm$1.12 & 54.38$\pm$0.47 & 58.21$\pm$0.74 & \textbf{60.53$\pm$0.31} \\
          & MF1   & 59.18$\pm$1.23  & 41.59$\pm$1.03 & 57.83$\pm$2.24 & \underline{59.20$\pm$0.67} & 53.10$\pm$1.92 & 54.29$\pm$0.48 & 57.68$\pm$1.02 & \textbf{60.47$\pm$0.51} \\
    \midrule
    \multirow{2}[2]{*}{IW} & Accu  & 44.24$\pm$0.31 &  30.19$\pm$0.27 & 56.08$\pm$1.22 & \underline{58.60$\pm$0.35} & 29.30$\pm$2.34 & 24.45$\pm$2.35 & 58.07$\pm$0.31 & \textbf{65.80$\pm$0.36} \\
          & MF1   & 43.67$\pm$0.30 &  28.86$\pm$1.02 & 55.72$\pm$1.22 & \underline{58.16$\pm$0.42} & 24.29$\pm$2.46 & 22.29$\pm$2.40 & 57.90$\pm$0.30 & \textbf{65.49$\pm$0.48} \\
    \midrule
    \multirow{2}[2]{*}{MI} & Accu  & \underline{54.00$\pm$0.50} & 52.40$\pm$3.12 & 51.70$\pm$4.63 & 53.00$\pm$0.49 & 52.45$\pm$0.78 & 51.00$\pm$1.67 & 50.00$\pm$2.42 & \textbf{56.00$\pm$0.46} \\
          & MF1   & \textbf{53.98$\pm$0.49} &  \underline{52.59$\pm$3.85} & 46.53$\pm$5.88 & 48.87$\pm$0.58 & 38.64$\pm$1.12 & 17.54$\pm$2.45 & 46.94$\pm$3.21 & 50.25$\pm$0.36 \\
    \midrule
    \multirow{2}[2]{*}{SRSCP1} & Accu  & 80.99$\pm$0.91 &  76.76$\pm$1.27 & 83.64$\pm$0.99 & 82.94$\pm$1.67 & 82.31$\pm$1.01 & \underline{83.95$\pm$1.15} & 82.18$\pm$1.10 & \textbf{84.47$\pm$1.19} \\
          & MF1   & 80.97$\pm$0.92 &  75.90$\pm$1.02 & 83.61$\pm$0.99 & 82.92$\pm$1.68 & 81.81$\pm$1.00 & \underline{83.81$\pm$1.15} & 82.18$\pm$1.10 & \textbf{84.44$\pm$1.19} \\
    \bottomrule
    \bottomrule
    \end{tabular}%
  \caption{Comparisons with State-of-the-Art methods for different tasks (\%)}
  \label{tab:sota}%
\end{table*}%

\subsection{Ablation Study}

\begin{table*}[htbp]
  \centering
  \small
    \begin{tabular}{cl|ll|lll|l}
    \toprule
    \toprule
    \multicolumn{2}{c|}{\multirow{2}[2]{*}{TS-GAC (Variants)}} & \multicolumn{2}{c|}{Augmentations} & \multicolumn{3}{c|}{Contrasting} & \multicolumn{1}{c}{\multirow{2}[2]{*}{Complete}} \\
    \multicolumn{2}{c|}{} & w/o Aug. (N) & w/o Aug. (E)& w/o GC & w/o NC & w/o MWTC  &  \\
    \midrule
    \multirow{2}[2]{*}{HAR} & Accu  & 92.97$\pm$0.23 & 93.67$\pm$0.11& 92.10$\pm$0.09 & 92.29$\pm$0.27 & 93.60$\pm$0.19  & 94.27$\pm$0.12 \\
          & MF1   & 92.69$\pm$0.27 & 93.41$\pm$0.12& 91.76$\pm$0.11 & 92.03$\pm$0.30 & 93.38$\pm$0.21  & 94.07$\pm$0.14 \\
    \midrule
    \multirow{2}[2]{*}{ISRUC} & Accu  & 83.86$\pm$0.20 & 83.87$\pm$0.18& 81.70$\pm$0.14 & 81.29$\pm$0.12 & 81.29$\pm$0.34  & 84.22$\pm$0.17 \\
          & MF1   & 82.88$\pm$0.28 & 82.80$\pm$0.15 & 80.62$\pm$0.13 & 80.11$\pm$0.11 & 80.11$\pm$0.87 & 83.45$\pm$0.23 \\
    \bottomrule
    \bottomrule
    \end{tabular}%
  \caption{Ablation study for graph augmentation and graph contrasting (\%)}
  \label{tab:abla}%
\end{table*}%

We evaluate designed augmentation and contrasting techniques within TS-GAC, which fall into two categories of variants. The first category tests augmentations, including w/o Aug. (N) and w/o Aug. (E), representing variants without node and edge augmentations, respectively. The second category assesses the effectiveness of contrastive losses, with variants w/o GC, w/o NC, and w/o MWTC indicating the removal of graph-level contrasting, node-level contrasting, and multi-window temporal contrasting, respectively. Finally, we compare them with the complete TS-GAC.

Table \ref{tab:abla} shows the results, where we only present the
results on HAR and ISRUC due to limited space. 
\textit{More results can be found in our supplementary materials}. The experimental results demonstrate the effectiveness of our proposed graph augmentation and contrasting techniques in achieving spatial consistency for MTS data. Specifically, the graph augmentations show significant improvements in learning robust representations. Compared to the variant without node augmentations, our complete TS-GAC achieves improvements of 1.30\% and 0.36\% on the two datasets. Similarly, compared to the model without edge augmentations, our complete TS-GAC achieves improvements of 0.60\% and 0.35\% on the two datasets. The improvements indicate the necessity of using graph augmentations for better augmenting MTS data. Meanwhile, the designed contrasting techniques play crucial roles in learning robust representations, and our complete TS-GAC achieves the best performance compared to the variants without any of the contrastive losses. For instance, we see drops of 2.17\% and 2.52\% by removing GC and drops of 1.98\% and 2.93\% by removing NC on the two datasets, indicating the effectiveness of graph contrasting in achieving spatial consistency. We further observe drops of 0.67\% and 2.90\% by removing MWTC on the two datasets, showing the importance of achieving temporal consistency for each sensor. Additionally, we can derive from the results that TS-GAC can still achieve good performance even when only graph contrasting is used, further highlighting the effectiveness of graph contrasting.

Overall, these findings validate the importance of our proposed graph augmentation and contrasting techniques, demonstrating the necessity of achieving spatial consistency when conducting CL for MTS data.

\subsection{Sensitivity Analysis}

\begin{figure}[htbp!]
    \centering
    \includegraphics[width = 1\linewidth]{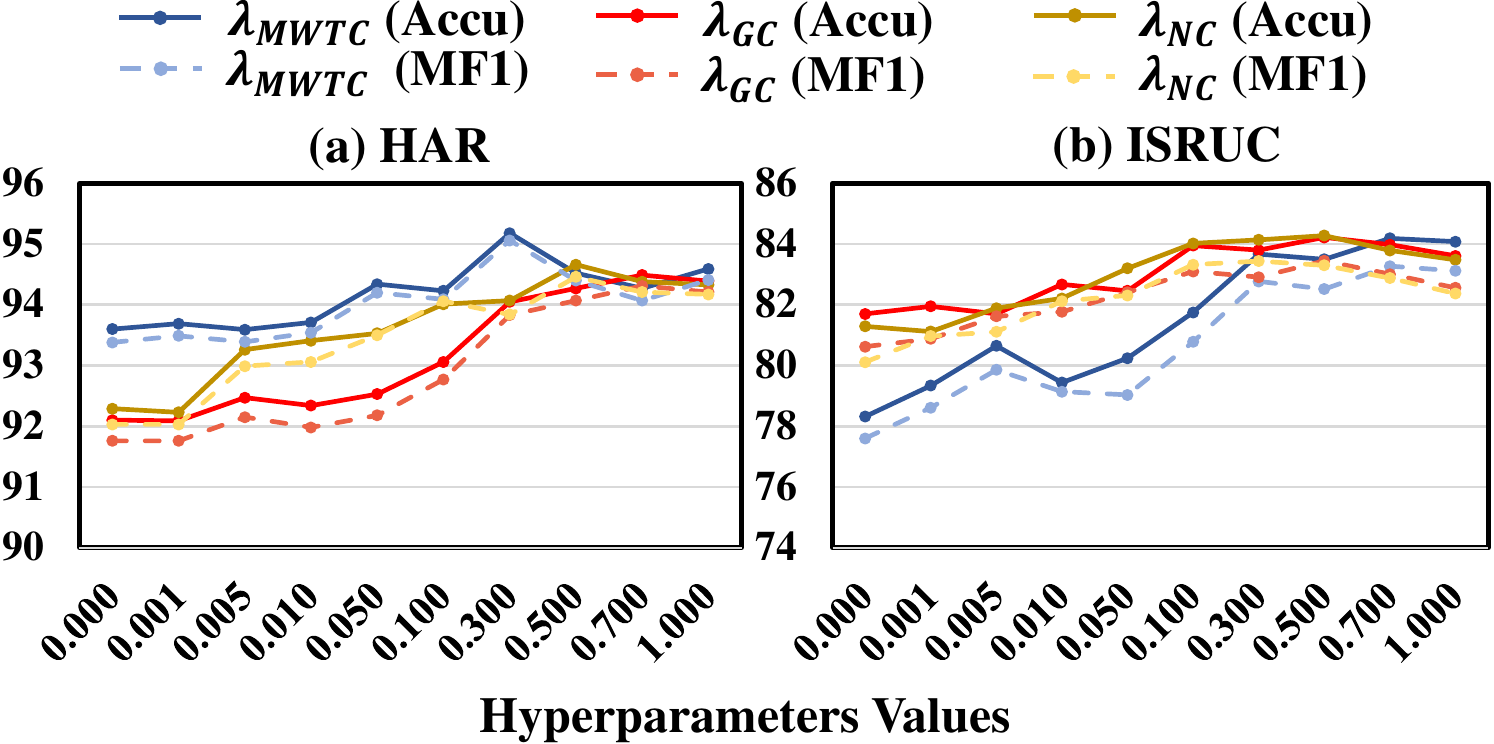}
    \caption{Sensitivity analysis for $\lambda_{MWTC}$, $\lambda_{GC}$, and $\lambda_{NC}$.}
    \label{fig:sensi_lambda}
\end{figure}
\paragraph{Hyperparameter Analysis}

We analyze $\lambda_{MWTC}$, $\lambda_{GC}$, and $\lambda_{NC}$ to test their effects. The hyperparameters are trade-offs between various losses, so we choose the values within [0, 0.001, 0.005, 0.01, 0.05, 0.1, 0.3, 0.5, 0.7, 1.0]. To reduce computation costs, we fixed other hyperparameters as 1 when testing one of them. From Fig.~\ref{fig:sensi_lambda}, we observe that TS-GAC tends to achieve better performance when the hyperparameters are set as larger values. For example, the accuracy increases by 2\% with $\lambda_{GC}$ increasing from 0 to 0.7 on HAR. The improvements show that the contrastive losses have positive effects on CL performance. However, the improvements become small when the values are large enough. For example, the performance has no obvious improvements with increasing $\lambda_{GC}$ from 0.7 to 1. Similar trends can also be found in other hyperparameters. Therefore, we can derive that the large hyperparameters have positive effects on the performance; however, too large values are unnecessary.

\begin{figure}[htbp!]
    \centering
    \includegraphics[width = 1\linewidth]{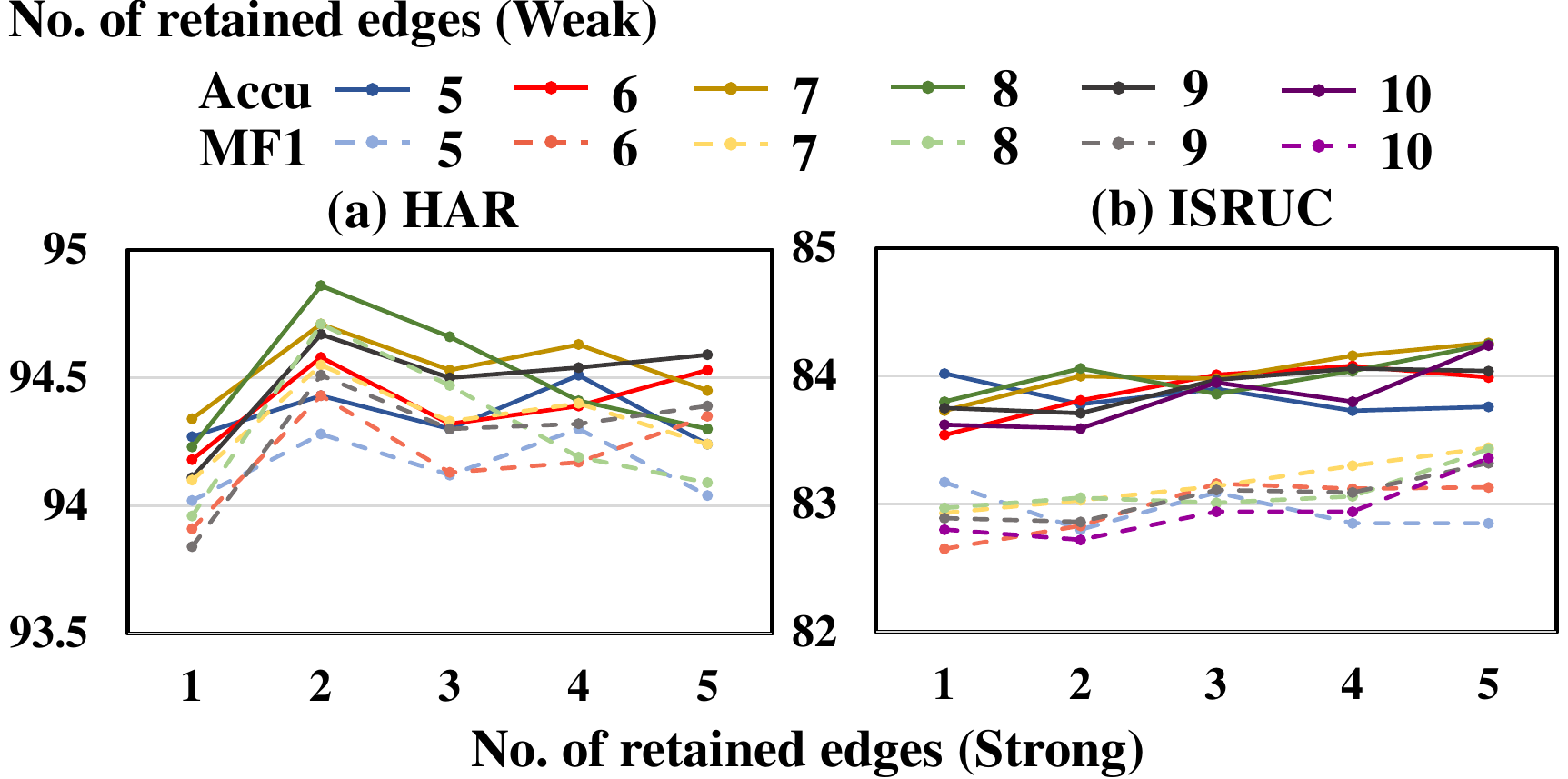}
    \caption{Sensitivity analysis for retained edges in views.}
    \label{fig:sensi_remained}
\end{figure}
\paragraph{Number of retained edges for edge augmentations}
To effectively augment sensor correlations, we design edge augmentations by retaining the $s$ strongest correlations, i.e., edges, for each sensor and replacing remaining correlations with random values. The value of $s$ is crucial for augmenting sensor relations and thus requires testing. Here, the weak view should have larger $s$ for weak augmentation while the strong view should have smaller $s$ for strong augmentation. Meanwhile, each sensor in HAR and ISRUC has 9 and 10 edges respectively. Thus, we set $s$ in the weak view within [5,6,7,8,9] for HAR and add 10 for ISRUC. For the strong view, we set $s$ within [1, 2, 3, 4, 5] for both datasets. Fig.~\ref{fig:sensi_remained} shows the results on HAR and ISRUC, where no. of retained edges represents the value of $s$. We take the results in HAR for example, and observe that our model shows better performance when $s$ in the strong view is set to 2 while keeping $s$ in the weak view fixed. On the other hand, our model shows better performance when $s$ in the weak view is set to 7 or 8 while keeping $s$ in the strong view fixed. These trends indicate that having fewer retained correlations in the strong view has a positive effect, but the value of $s$ should not be too small so as to avoid overly distorted correlations. Similarly, having more retained correlations in the weak view is beneficial, but the value of $s$ should not be too large. 


\section{Conclusion}

We propose TS-GAC for MTS classification. To achieve spatial consistency, specific augmentation and contrasting techniques are tailored for MTS data. To better augment MTS data, graph augmentations are proposed, including node and edge augmentations for ensuring the robustness of sensors and their correlations. Besides, graph contrasting is designed, including node- and graph-level contrasting to extract robust sensor- and global-level features. We further introduce multi-window temporal contrasting to ensure temporal consistency for each sensor. Experiments show that TS-GAC achieves SOTA performance in various MTS classification tasks.
\section*{Acknowledgements}
We thank anonymous reviewers for their constructive comments on this work. This research is supported by the Agency for Science, Technology and Research (A*STAR) under its AME Programmatic Funds (Grant No. A20H6b0151) and Career Development Award (Grant No. C210112046), and the National Research Foundation, Singapore under its AI Singapore Programme (AISG2-RP-2021-027).
\bibliography{aaai24}

\end{document}